\documentclass[runningheads]{llncs}
\usepackage[T1]{fontenc}
\usepackage{graphicx}
\usepackage{todonotes}
\usepackage{subcaption}
\usepackage{subcaption}
\usepackage{hyperref}       
\usepackage{url}            
\usepackage{booktabs}       
\usepackage{amsfonts}       
\usepackage{nicefrac}       
\usepackage{microtype}      
\usepackage{color}

\begin{document}
\title{Deployment of Image Analysis Algorithms \\ under Prevalence Shifts}
%
%
\author{Patrick Godau\inst{1,2,3,4,\star} \and 
Piotr Kalinowski\inst{1,4,\star} \and Evangelia Christodoulou\inst{1} \and Annika Reinke\inst{1,3,5} \and Minu Tizabi\inst{1,5} \and Luciana Ferrer\inst{6} \and Paul Jäger\inst{5,7} \and 
 Lena Maier-Hein\inst{1,2,3,5,8}}

\renewcommand{\thefootnote}{\fnsymbol{footnote}}
\footnotetext[1]{P. Godau and P. Kalinowski contributed  equally to this paper.}

 \authorrunning{P. Godau, P. Kalinowski et al.}
%
 \institute{Division of Intelligent Medical Systems (IMSY), German Cancer Research Center (DKFZ), Heidelberg, Germany 
 \email{patrick.godau@dkfz-heidelberg.de} \and
 National Center for Tumor Diseases (NCT), NCT Heidelberg, a partnership between DKFZ and university medical center Heidelberg 
 \and Faculty  of  Mathematics  and  Computer  Science, Heidelberg University, Germany
\and
 HIDSS4Health - Helmholtz Information and Data Science School for Health, Karlsruhe/Heidelberg, Germany 
 \and
  Helmholtz Imaging,  German  Cancer  Research  Center  (DKFZ),  Germany
\and Instituto de Ciencias de la Computación, UBA-CONICET, Argentina \and Interactive Machine Learning Group, German Cancer Research Center (DKFZ), Germany 
 \and
Medical  Faculty, Heidelberg University, Germany}

\maketitle              

\begin{abstract}
Domain gaps are among the most relevant roadblocks in the clinical translation of machine learning (ML)-based solutions for medical image analysis. While current research focuses on new training paradigms and network architectures, little attention is given to the specific effect of prevalence shifts on an algorithm deployed in practice. Such discrepancies between class frequencies in the data used for a method's development/validation and that in its deployment environment(s) are of great importance, for example in the context of artificial intelligence (AI) democratization, as disease prevalences may vary widely across time and location. Our contribution is twofold. First, we empirically demonstrate the potentially severe consequences of missing prevalence handling by analyzing (i) the extent of miscalibration, (ii) the deviation of the decision threshold from the optimum, and (iii) the ability of validation metrics to reflect neural network performance on the deployment population as a function of the discrepancy between development and deployment prevalence. Second, we propose a workflow for prevalence-aware image classification that uses estimated deployment prevalences to adjust a trained classifier to a new environment, without requiring additional annotated deployment data. Comprehensive experiments based on a diverse set of 30 medical classification tasks showcase the benefit of the proposed workflow in generating better classifier decisions and more reliable performance estimates compared to current practice.

\end{abstract}

\section{Introduction}\label{sec:intro}

\begin{figure}[!t]
  \centering
  \includegraphics[width=0.9\linewidth]{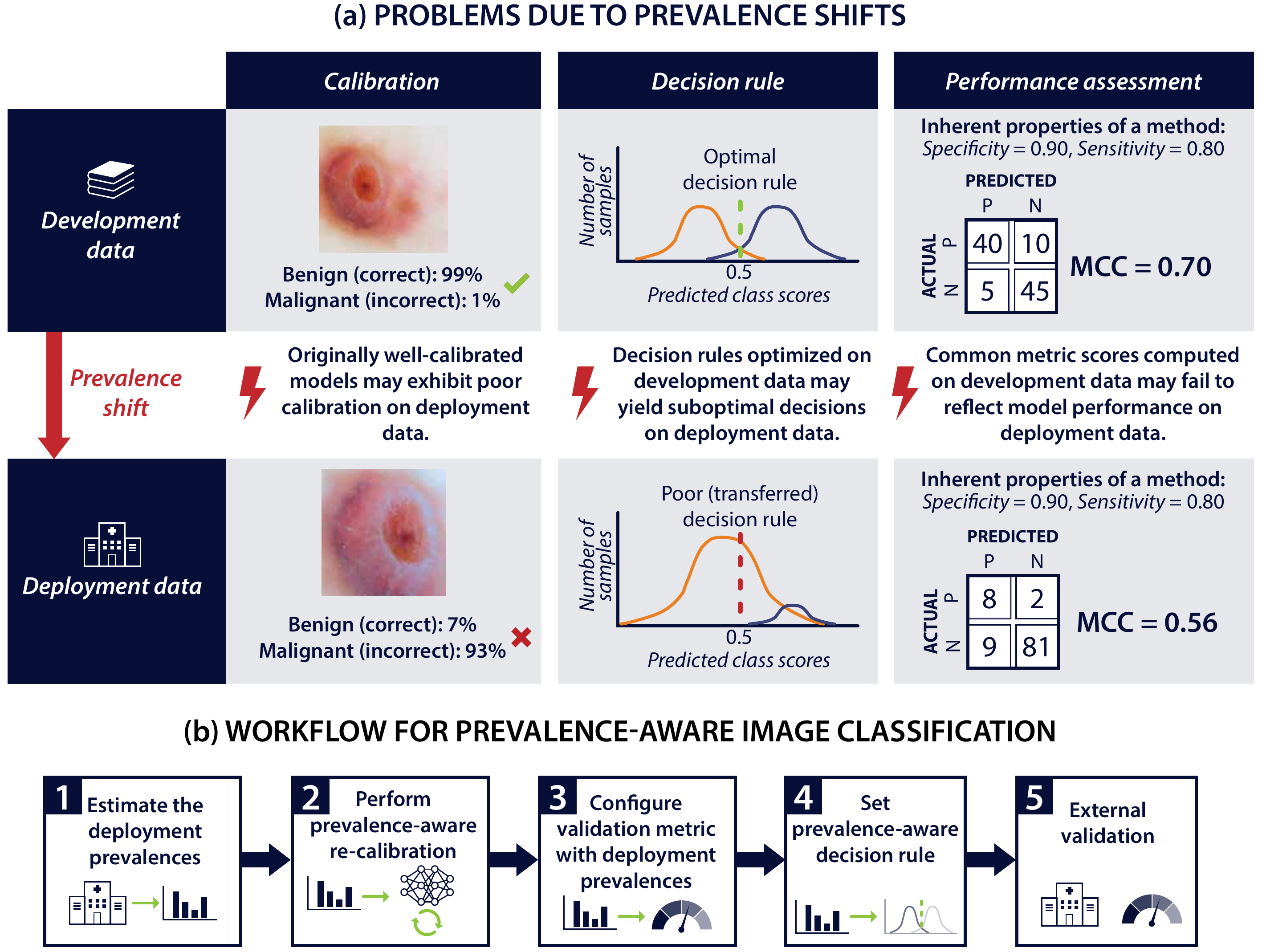}
  \caption{Summary of contributions. (a) Based on a dataset comprising 30 medical image classification tasks, we show that prevalence shifts between development data and deployment data engender various problems. (b) Our workflow for prevalence-aware medical image classification addresses all of these issues.
}
  \label{fig:concept}
\end{figure}

Machine learning (ML) has begun revolutionizing many fields of imaging research and practice. The field of medical image analysis, however, suffers from a substantial translational gap that sees a large number of methodological developments fail to reach (clinical) practice and thus stay short of generating (patient) benefit. A major roadblock are dataset shifts, situations in which the distributions of data used for algorithm development/validation and its deployment, differ due to exogenous factors such as dissimilar cohorts or differences in the acquisition process \cite{Castro2019CausalityMI,Zhang2022ShiftingML}. In the following, we focus on prevalence shifts, 
which are highly relevant in the context of 
global artificial intelligence (AI) \cite{Subbaswamy2019FromDT}. Common causes for prevalence shifts include sample selection bias and variations in environmental factors like season or geography \cite{Castro2019CausalityMI,Dockes2021PreventingDS,Zhang2022ShiftingML}.
According to prior work \cite{Dockes2021PreventingDS} as well as our own analyses, prevalence handling is especially crucial in the following steps related to model deployment:

\textbf{Model re-calibration}: After a prevalence shift models need to be re-calibrated. This has important implications on the decisions made based on predicted class scores (see next point). Note in this context that deep neural networks tend not to be calibrated after training in the first place \cite{Guo2017OnCO}.

\textbf{Decision rule}: A decision rule is a strategy transforming continuous predicted class scores into a single classification decision. Simply using the argmax operator ignores the theoretical boundary conditions derived from Bayes theory. Importantly, argmax relies on the predicted class scores to be calibrated and is thus highly sensitive to prevalence shifts \cite{Ferrer2022AnalysisAC}. Furthermore, it only yields the optimal decision for specific metrics. Analogously, tuned decision rules may not be invariant to prevalence shifts. 

\textbf{Performance assessment}: Class frequencies observed in one test set are in general not representative of those encountered in practice. This implies that the scores for widely used prevalence-dependent metrics, such as Accuracy, F1 Score, and Matthews Correlation Coefficient (MCC), would substantially differ when assessed under the prevalence shift towards clinical practice \cite{MaierHein2022MetricsRP}. 

This importance, however, is not reflected in common image analysis practice. Through a literature analysis, we found that out of a total of 53 research works published between 01/2020 and beginning of 03/2023 that used any of the data included in our study, only one explicitly mentioned re-calibration. Regarding the most frequently implemented decision rules, roughly three quarters of publications did not report any strategy, which we strongly assume to imply use of the default argmax operator. Moreover, both our analysis and previous work show Accuracy and F1 Score to be among the most frequently used metrics for assessing classification performance in comparative medical image analysis \cite{MaierHein2018WhyRO,MaierHein2022MetricsRP}, indicating that severe performance deviations under potential prevalence shifts are a widespread threat.

Striving to bridge the translational gap in AI-based medical imaging research caused by prevalence shifts, our work provides two main contributions:
First, we demonstrate the potential consequences of ignoring prevalence shifts on a diverse set of medical classification tasks. Second, we assemble a comprehensive workflow for image classification, which is robust to prevalence shifts. As a key advantage, our proposal requires only an estimate of the expected prevalences rather than annotated deployment data and can be applied to any given black box model.

\begin{figure}[t]
  \centering
  \includegraphics[width=\linewidth]{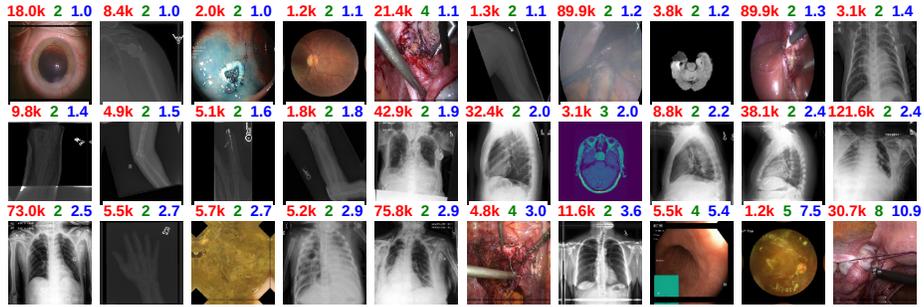}
  \caption{Medical image classification tasks used in this study. The number of samples (red) and classes (green) ranges from 1,200 to 121,583 and two to eight, respectively. The imbalance ratio (blue) varies between 1 and 10.9.}
  \label{fig:samples}
\end{figure}

\section{Methods}\label{sec:methods}

\subsection{Workflow for prevalence-aware image classification}

Our workflow combines existing components of validation in a novel manner. As illustrated in Fig. \ref{fig:concept}, it leverages estimated deployment prevalences to adjust an already trained model to a new environment. We use the following terminology. 

\textbf{Fundamentals}: We define a dataset $D := \{(x_i, y_i) | 1 \leq i \leq N\}$ by a set of $N$ images $x_i \in X$ and labels $y_i \in Y$ with $Y = \{1, \ldots, C\}$. $P_D$ is a $C$-dimensional vector called the prevalences of $D$, where $P_D(k): = |\{(x_i, y_i) \in D | y_i = k\}| / N$ is the prevalence of class $k \in Y$. 
The fraction $\max_k \{P_D(k)\} / \min_k \{P_D(k)\}$ is named the imbalance ratio (IR) of $D$. 

\textbf{Re-calibration}: We refer to the output of a model $\varphi: X \rightarrow \mathbb{R}^C$ before applying the softmax activation as $\varphi(x)$. It can be re-calibrated by applying a transformation $f$. Taking the softmax of $\varphi(x)$ (no re-calibration) or of $f(\varphi(x))$, we obtain predicted class scores $s_x$. The probably most popular re-calibration approach is referred to as “temperature scaling” \cite{Guo2017OnCO} and requires only a single parameter $t \in \mathbb{R}$ to be estimated: $f_\mathrm{temp}(\varphi(x)) = \varphi(x) / t$. 
The transformation parameter(s) is/are learned with minimization of the cross-entropy loss. 

\textbf{Decision rule}: A decision rule $d$ is a deterministic algorithm that maps predicted class scores $s_x$ to a final prediction $d(s_x) \in Y$. 
The most widely used decision rule is the argmax operator, although various alternatives exist \cite{MaierHein2022MetricsRP}.

To overcome problems caused by prevalence shifts, we propose the following workflow (Fig. \ref{fig:concept}b)

\textbf{Step 1: Estimate the deployment prevalences}: The first step is to estimate the prevalences in the deployment data $D_{dep}$, e.g., based on medical records, epidemiological research, or a data-driven approach \cite{Lipton2018DetectingAC,Saerens2002AdjustingTO}. 
The workflow requires an underlying anticausal connection of image and label, i.e., a label $y$ causes the image $x$ (e.g., presence of a disease has a visual effect)\cite{Castro2019CausalityMI,Dockes2021PreventingDS}, to be verified at this point.  

\textbf{Step 2: Perform prevalence-aware re-calibration}: 
Given an external factor that varies $P_D$ between development calibration and deployment datasets $D_{cal}$ and $D_{dep}$ from, we can assume the likelihoods $P(x | y = k)$ to stay identical for an anticausal problem, ignoring potential manifestation and acquisition shifts during deployment \cite{Castro2019CausalityMI}. Under mild assumptions \cite{Lipton2018DetectingAC,Zhang2013DomainAU}, weight adaptation in the loss function optimally solves the prevalence shift for a classifier. In the presence of prevalence shifts, we therefore argue for adaptation of weights in the cross-entropy loss $\sum_i -w(y_i) \log(s_i(y_i))$ according to the expected prevalences; more precisely, for class $k$ we use the weight $w(k) = P_{D_{dep}}(k) / P_{D_{cal}}(k)$ during the learning of the transformation parameters \cite{Dockes2021PreventingDS,Shimodaira2000ImprovingPI,Zhang2013DomainAU}. Furthermore, since temperature scaling's single parameter $t$ is incapable of correcting the shift produced by a mismatch in prevalences, we add a bias term $b \in \mathbb{R}^C$ to be estimated alongside $t$ as suggested by \cite{Alexandari2020MaximumLW,Brummer2006OnCO,Platt1999ProbabilisticOF}. We refer to this re-calibration approach as “affine scaling”:
$f_\mathrm{aff}(\varphi(x)) = \varphi(x) / t + b$. 

\textbf{Step 3: Configure validation metric with deployment prevalences}: Prevalence-dependent metrics, such as Accuracy, MCC, or the F1 Score, are widely used in image analysis due to their many advantages \cite{MaierHein2022MetricsRP}. However, they reflect a model's performance only with respect to the specific, currently given prevalence. This problem can be overcome with the metric Expected Cost (EC) \cite{Ferrer2022AnalysisAC}. In its most general form, we can express EC as $\mathrm{EC} = \sum_k P_D(k) \sum_j c_{kj} R_{kj}$, where $c_{kj}$ refers to the “costs” we assign to the decision of classifying a sample of class $k$ as $j$ and $R_{kj}$ is the fraction of all samples with reference class $k$ that have been predicted as $j$. Note that the standard 0-1 costs ($c_{kk} = 0$ for all $k$ and $c_{kj} = 1$ for $k \neq j$) reduces to EC being 1 minus Accuracy. To use EC as a robust estimator of performance, we propose replacing the prevalences $P_D(k)$ with those previously estimated in step 1 \cite{Ferrer2022AnalysisAC}. 

\textbf{Step 4: Set prevalence-aware decision rule}: Most counting metrics \cite{MaierHein2022MetricsRP} require some tuning of the decision rule during model development, as the argmax operator is generally not the optimal option. This tuning relies on data from the development phase and the resulting decision rule is likely dependent on development prevalences and does not generalize (see Sec. \ref{sec:results}). On the other hand, EC, as long as the predicted class scores are calibrated, yields the optimal decision rule $\mathrm{argmin}_k \sum_j c_{jk} s_x(j)$  \cite{Bishop2006PatternRA,Hastie2001TheEO}. For standard 0-1 costs, this simplifies to the argmax operator.

\textbf{Step 5: External validation}: The proposed steps for prevalence-aware image classification have strong theoretical guarantees, but additional validation on the actual data of the new environment is indispensable for monitoring \cite{Saria2019TutorialSA}.

\subsection{Experimental design}\label{ss:experimental}
The purpose of our experiments was twofold: (1) to quantify the effect of ignoring prevalence shifts when validating and deploying models and (2) to show the value of the proposed workflow. The code for our experiments is available at \url{https://github.com/IMSY-DKFZ/prevalence-shifts}.

\subsubsection{Medical image classification tasks}\label{ss:data}

To gather a wide range of image classification tasks for our study, we identified medical image analysis tasks that are publicly available and provide at least 1000 samples. This resulted in 30 tasks covering the modalities laparoscopy \cite{Leibetseder18,Twinanda2017EndoNetAD}, gastroscopy/colonoscopy \cite{borgli2020,Pogorelov2017}, magnetic resonance imaging (MRI)\cite{jakeshbohaju_2020,cheng_2017}, X-ray \cite{HematejaAluru2021,irvin2019chexpert,KERMANY20181122,Pranav2017}, fundus photography \cite{LIU2022100512}, capsule endoscopy\cite{Smedsrud2021}, and microscopy \cite{Negin2020} (Fig. \ref{fig:samples}).
We split each task as follows: 30\% of the data -- referred to as “deployment test set” $D_{dep}$ -- was used as a hold-out split to sample subsets $D_{dep}(r)$ representing a deployment scenario with IR r. The remaining data set made up the “development data“, comprising the “development test set” $D_{test}$  (10\%; class-balanced) , the “training set” (50\%) and the “validation set” (10\%; also used for calibration). 

\subsubsection{Experiments}
For all experiments, the same neural network models served as the basis. To mimic a prevalence shift, we sub-sampled datasets $D_{dep}(r)$ from the deployment test sets $D_{dep}$ according to IRs $r \in [1, 10]$ with a step size of $0.5$. The experiments were performed with the popular prevalence-dependent metrics Accuracy, MCC, and F1 Score, as the well as EC with 0-1 costs.
For our empirical analyses, we trained neural networks (specifications: see Tab. \ref{tab:models} Suppl.) for all 30 classification tasks introduced in Sec. \ref{ss:data}. In the interest of better reproducibility and interpretability, we focused on a homogeneous workflow (e.g., by fixing hyperparameters across tasks) rather than aiming to achieve the best possible Accuracy for each individual task. The following three experiments were performed. 
(1) To assess the \textbf{effects of prevalence shifts on model calibration}, we measured miscalibration on the deployment test set $D_{dep}(r)$ as a function of the increasing IR r for five scenarios: no re-calibration, temperature scaling, and affine scaling (the latter two with and without weight adaptation). 
Furthermore, (2) to assess the \textbf{effects of prevalence shifts on the decision rule}, for the 24 binary tasks, we computed -- with and without re-calibration and for varying IR r -- the differences between the metric scores on $D_{dep}(r)$ corresponding to an \textit{optimal} decision rule and two other decision rules: argmax and a cutoff that was tuned on $D_{test}$.
Lastly, (3) to assess the \textbf{effects of prevalence shifts on the generalizability of validation results}, we measured the absolute difference between the metric scores obtained on the development test data $D_{test}$ and those obtained  on the deployment test data $D_{dep}(r)$ with varying IR r. The scores were computed for the argmax decision rule for both non-re-calibrated and re-calibrated predicted class scores.
To account for potential uncertainty in estimating deployment prevalences, we repeated all experiments with slight perturbation of the true prevalences. To this end, we drew the prevalence for each class from a normal distribution with a mean equal to the real class prevalence and fixed standard deviation (std). We then set a minimal score of 0.01 for each class and normalized the resulting distribution.

\section{Results}\label{sec:results}
\begin{figure}[t]
  \centering
  \includegraphics[width=0.95\linewidth]{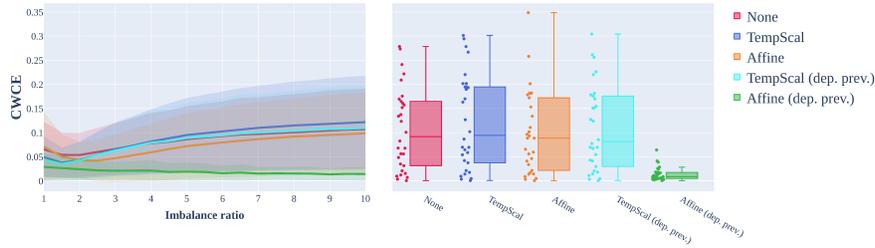}
  \caption{Effect of prevalence shifts on the calibration. The class-wise calibration error (CWCE) generally increases with an increasing prevalence shift from development (balanced) to deployment test set. Left: Mean (line) and standard deviation (shaded area) obtained from n = 30 medical classification tasks. Right: CWCE values for all tasks at imbalance ratio 10.
}
  \label{fig:calibration}
\end{figure}

\begin{figure}[t]
  \centering
  \includegraphics[width=0.95\linewidth]{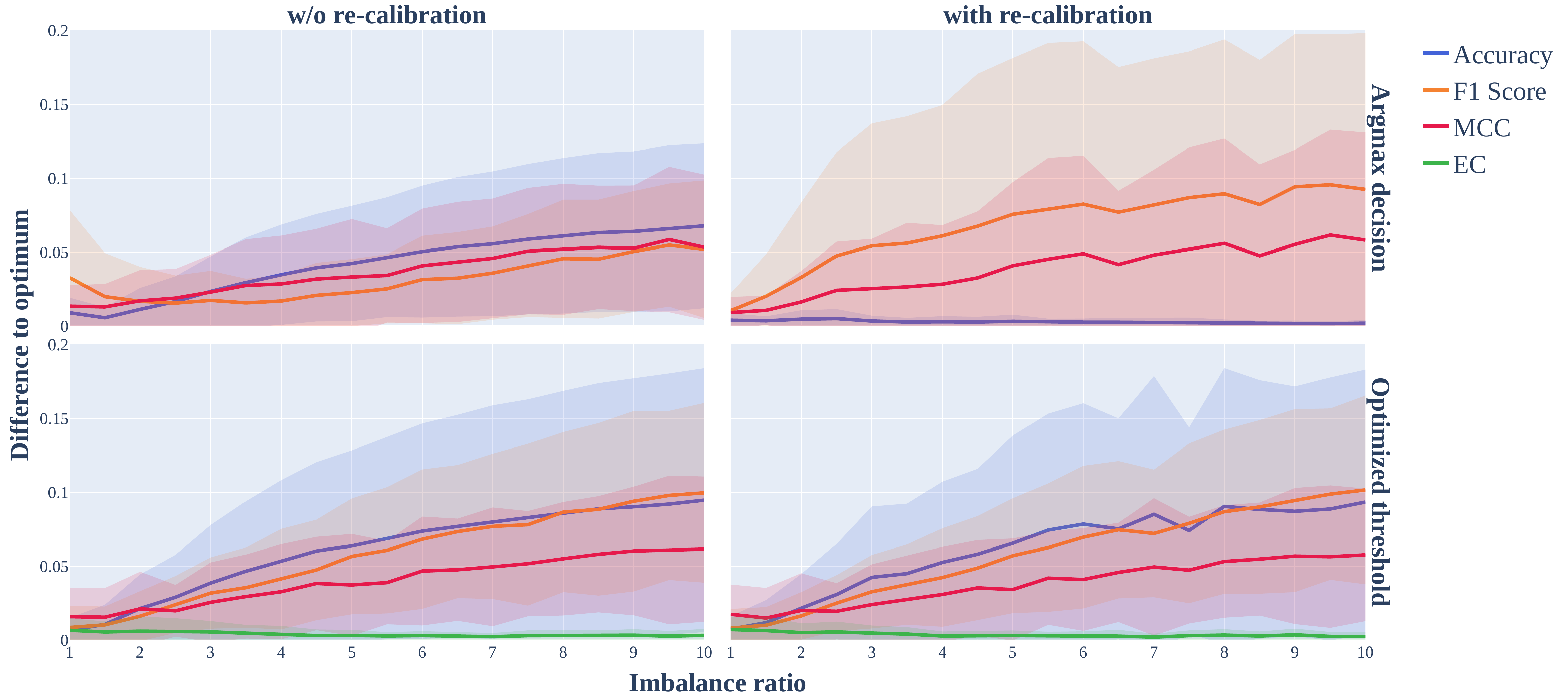}
  \caption{Effect of prevalence shifts on the decision rule. The difference between the actual metric score and the optimal metric score (optimal decision rule) is shown as a function of the imbalance ratio for non-re-calibrated (left) and re-calibrated (right) models for two decision rule strategies: argmax (top) and threshold optimization on the development test set (bottom). Mean (lines) and standard deviation (transparent area) obtained from n=24 binary tasks.
}
  \label{fig:threshold}
\end{figure}

\begin{figure}[!h]
  \centering
  \includegraphics[width=0.95\linewidth]{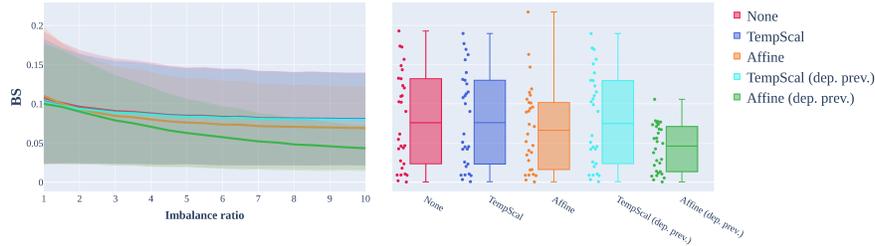}
  \caption{Effect of prevalence shifts on the  Brier Score (BS) in analogy to Fig. \ref{fig:calibration}.
  Our proposed affine recalibration with prevalences of the deployment data consistently provides the lowest BS.
}
  \label{fig:sup:calibration_metrics}
\end{figure}

\begin{figure}[!h]
  \centering
  \includegraphics[width=0.95\linewidth]{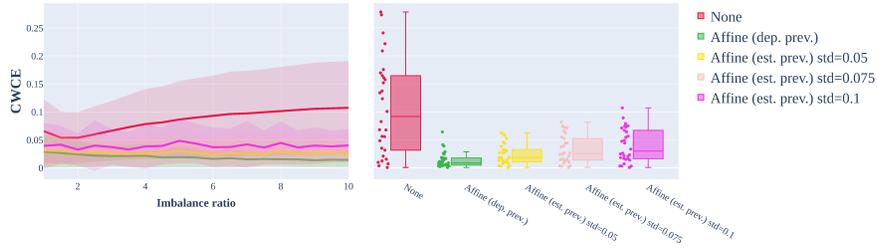}
  \caption{Effect of errors in the estimation of deployment prevalences on the performance of the proposed affine re-calibration method. In analogy to Fig. \ref{fig:calibration} the class-wise calibration error (CWCE) is plotted for n = 30 different biomedical imaging tasks for various prevalence shifts. 
}
  \label{fig:sup:calibration_perturb}
\end{figure}

\textbf{Effects of prevalence shifts on model calibration}
In general, the calibration error increases with an increasing discrepancy between the class prevalences in the development and the deployment setting (Fig. \ref{fig:calibration}). The results clearly demonstrate that a simple accuracy-preserving temperature scaling-based method is not sufficient under prevalence shifts. Only our proposed method, which combines an affine transformation with a prevalence-driven weight adjustment, consistently features good calibration performance. This also holds true when perturbing the deployment prevalences, as demonstrated in Fig. \ref{fig:sup:calibration_perturb} (Suppl.). For the inspected range (up to r=10), miscalibration can be kept constantly close to 0. Note that CWCE is a biased estimator of the canonical calibration error \cite{MaierHein2022MetricsRP}, which is why we additionally report the Brier Score (BS) as an overall performance measure (Fig. \ref{fig:sup:calibration_metrics} Suppl.). \newline

\begin{figure}[!h]
  \centering
  \includegraphics[width=0.95\linewidth]{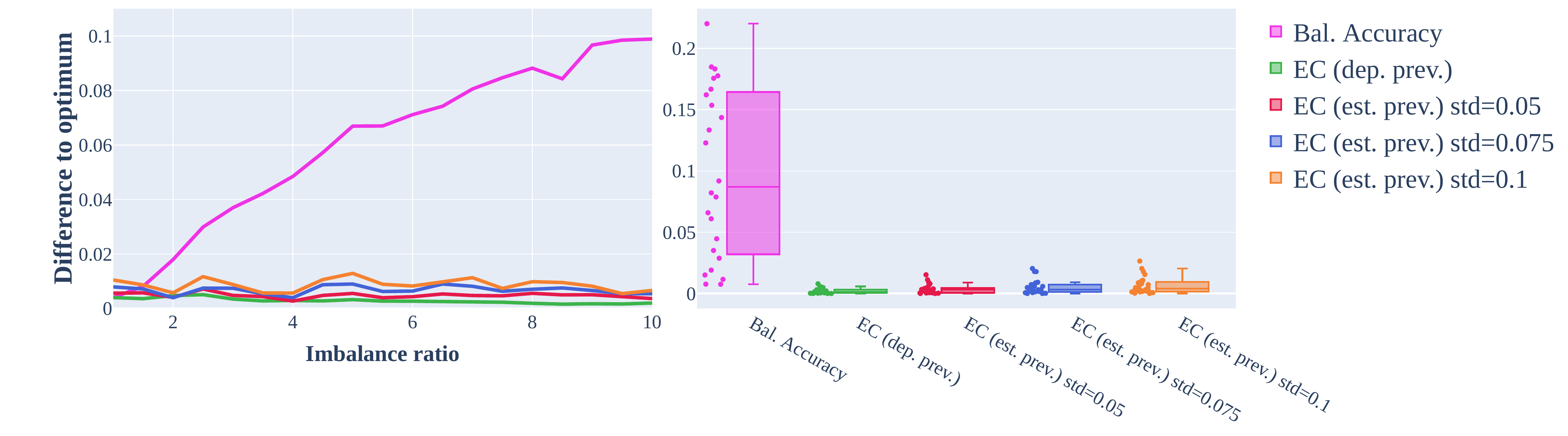}
  \caption{Effect of errors in the estimation of deployment prevalences on the quality of the decision rule. Analogously to Fig. \ref{fig:threshold} (top right), the difference between the actual and the optimal metric score is shown as a function of the imbalance ratio (IR) for a re-calibrated model with argmax decision rule. The boxplots show results for the 24 binary tasks at IR=10. Argmax remains close to being the optimal threshold for EC when non-exact priors are used in re-calibration. We show the same plots for the prevalence-independent Balanced Accuracy, yielding suboptimal performance under prevalence shifts when using argmax decision rule. 
}
  \label{fig:sup:threshold}
\end{figure}

\textbf{Effects of prevalence shifts on the decision rule}
Fig. \ref{fig:threshold} supports our proposal: An argmax-based decision informed by calibrated predicted class scores (top right) and assessed with the Expected Cost (EC) metric (identical to the blue Accuracy line in this case) yields optimal results irrespective of prevalence shifts. In fact, this approach substantially increases the quality of the decisions when compared to a baseline without re-calibration, as indicated by an average relative decrease of EC by 25\%. This holds true in a similar fashion for perturbed versions of the re-calibration (Fig. \ref{fig:sup:threshold} Suppl.). The results further show that argmax is not the best decision rule for F1 Score and MCC (Fig. \ref{fig:threshold} top). Importantly, decision rules optimized on a development dataset do not generalize to unseen data under prevalence shifts (Fig. \ref{fig:threshold} bottom). \newline

\begin{figure}[!t]
  \centering
  \includegraphics[width=0.87\linewidth]{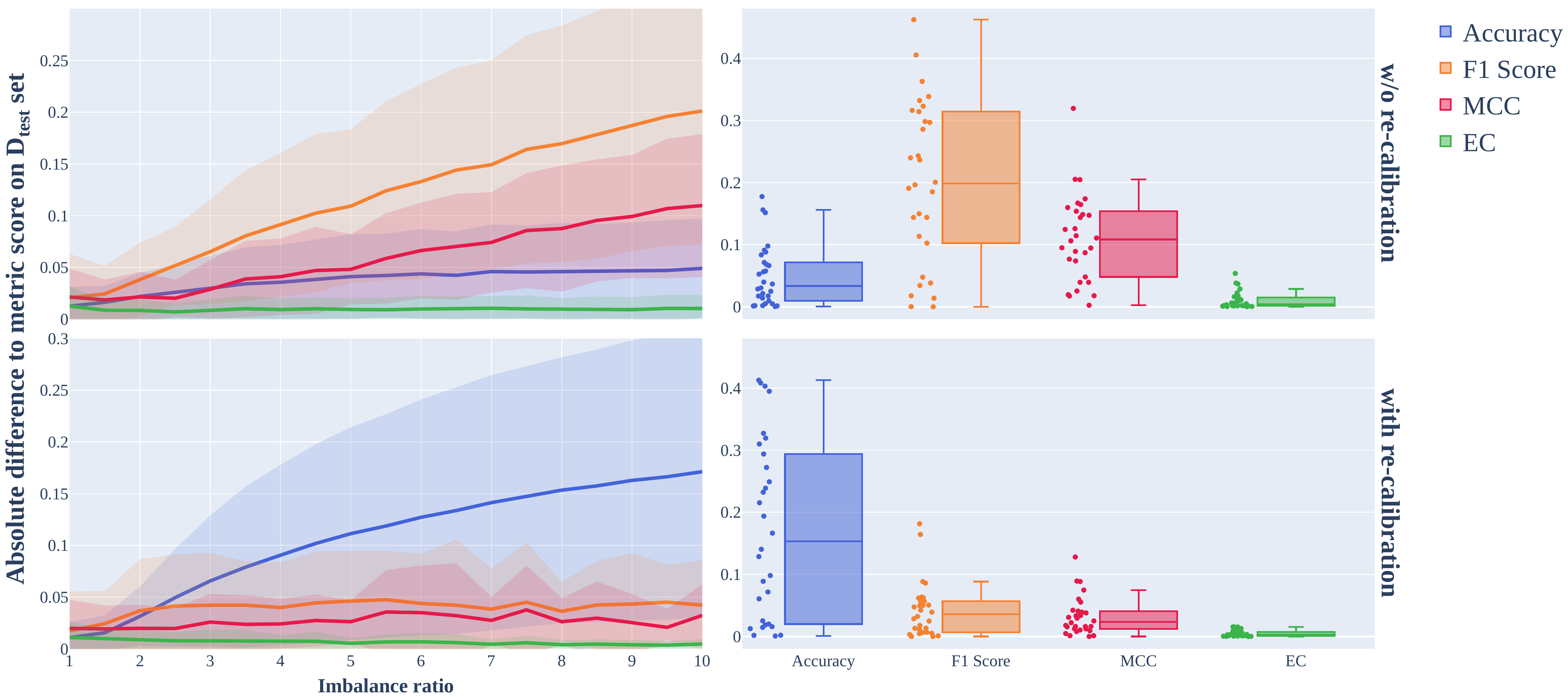}
  \caption{Effect of prevalence shifts on the generalizability of validation results. The absolute difference of the metric score computed on the deployment data to that computed on the development test set is shown as a function of the imbalance ratio (IR) for non-re-calibrated (top) and re-calibrated (bottom) models. The dot- and boxplots show the results for all n=30 tasks at a fixed IR of 10. 
}
  \label{fig:assessment}
\end{figure}

\begin{figure}[h]
  \centering
  \includegraphics[width=0.95\linewidth]{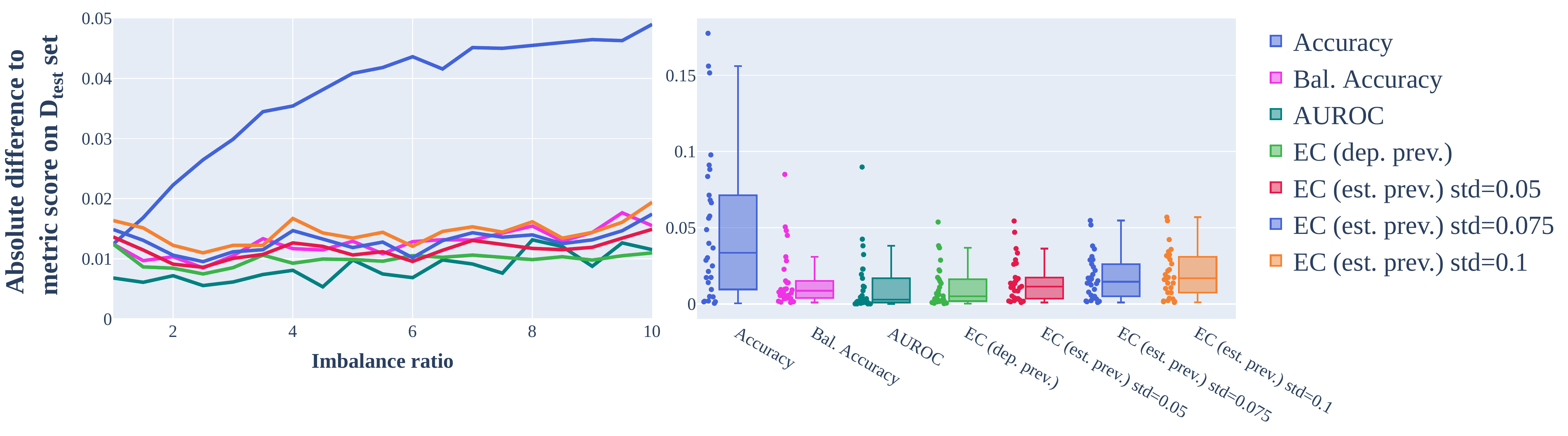}
  \caption{Effect of errors in the estimation of deployment prevalences on the generalizability of validation results when using Expected Cost (EC). Even under high prevalence shifts, the results remain stable. We further complement Fig. \ref{fig:assessment} (bottom) with the common prevalence-independent metrics Balanced Accuracy and AUROC, which also yield robust results.
}
  \label{fig:sup:assessment}
\end{figure}

\textbf{Effects of prevalence shifts on the generalizability of validation results}
As shown in Fig. \ref{fig:assessment}, large deviations from the metric score obtained on the development test data of up to 0.41/0.18 (Accuracy), 0.35/0.46 (F1 Score), and 0.27/0.32 (MCC), can be observed for the re-calibrated/non-re-calibrated case. In contrast, the proposed variation of Expected Cost (EC) enables a reliable estimation of performance irrespective of prevalence shifts, even when the prevalences are not known exactly (Fig. \ref{fig:sup:assessment} Suppl.). The same holds naturally true for the  prevalence-independent metrics Balanced Accuracy (BA) and Area under the Receiver Operating Curve (AUROC) (Fig. \ref{fig:sup:assessment} Suppl.).

\section{Discussion} 

Important findings, some of which are experimental confirmations of theory, are:
\begin{enumerate}
    \item Prevalence shifts lead to miscalibration. A weight-adjusted affine re-calibration based on estimated deployment prevalences compensates for this effect.
    \item Argmax should not be used indiscriminately as a decision rule. For the metric EC and specializations thereof (e.g., Accuracy), optimal decison rules may be derived from theory, provided that the predicted class scores are calibrated. This derived rule may coincide with argmax, but for other common metrics (F1 Score, MCC) argmax does not lead to optimal results. 
    \item An optimal decision rule, tuned on a development dataset, does not generalize to datasets with different prevalences. Prevalence-aware setting of the decision rule requires data-driven adjustment or selection of a metric with a Bayes theory-driven optimal decision rule.
    \item Common prevalence-dependent metrics, such as MCC and F1 Score, do not give robust estimations of performance under prevalence shifts. EC, with adjusted prevalences, can be used in these scenarios.
\end{enumerate}

These findings have been confirmed by repeated experiments using multiple random seeds for dataset splitting and model training. Overall, we present strong evidence that the so far uncommon metric EC offers key advantages over established metrics. Due to its strong theoretical foundation and flexibility in configuration it should, from our perspective, evolve to a default metric in image classification. Note in this context that while our study clearly demonstrates the advantages of prevalence-independent metrics, prevalence-dependent metrics can be much better suited to reflect the clinical interest \cite{MaierHein2022MetricsRP}. 

In conclusion, our results clearly demonstrate that ignoring potential prevalence shifts may lead to suboptimal decisions and poor performance assessment. In contrast to prior work \cite{Ma2022TesttimeAW}, our proposed workflow solely requires an estimation of the deployment prevalences -- and no actual deployment data or model modification. 
It is thus ideally suited for widespread adoption as a common practice in prevalence-aware image classification.

\subsubsection{Acknowledgements}
This project has been funded by (i) the German Federal Ministry of Health under the reference number 2520DAT0P1 as part of the pAItient (Protected Artificial Intelligence Innovation Environment for Patient Oriented Digital Health Solutions for developing, testing and evidence based evaluation of clinical value) project, (ii) HELMHOLTZ IMAGING, a platform of the Helmholtz Information \& Data Science Incubator and (iii) the Helmholtz Association under the joint research school “HIDSS4Health – Helmholtz Information and Data Science School for Health” (iv) state funds approved by the State Parliament of Baden-Württemberg for the Innovation Campus Health + Life Science Alliance Heidelberg Mannheim.

\bibliographystyle{splncs04}
\bibliography{references2990.bib}

\section{Supplementary Material}

\begin{table}[h]
\caption{Overview of additional results.}
\label{tab:overview}
\begin{tabular}{p{0.2\linewidth} p{0.8\linewidth}}

\toprule
Calibration & As CWCE is a biased estimator of calibration performance, we repeated the calibration analysis with the Brier Score (BS) in Fig. \ref{fig:sup:calibration_metrics}.\\
Robustness & Our workflow depends on the estimation of deployment prevalences, which may not be known exactly. We therefore repeated the analyses shown in Fig.~\ref{fig:calibration}-\ref{fig:assessment} with variations of the proposed method in which prevalences were perturbed. The method to generate  non-exact priors is described in Sec. \ref{ss:experimental}.\\
Prevalence-independent metrics & The metrics of the main paper feature specific advantages compared to prevalence-independent metrics, such as AUROC and BA~\cite{MaierHein2022MetricsRP} but come with the disadvantage of prevalence dependency. To investigate the performance of the proposed prevalence-corrected metric EC relative to common prevalence-independent metrics, we repeated the experiments  shown in Fig.~\ref{fig:threshold} and Fig.~\ref{fig:assessment} with BA and AUROC in Fig. \ref{fig:sup:threshold} and Fig. \ref{fig:sup:assessment}.\\
\bottomrule
\end{tabular}
\end{table}

\begin{table}[h]
\caption{Model training specifications.}
\label{tab:models}
\begin{tabular}{p{2.5cm}l}

\toprule
Architecture & ImageNet pretrained ResNet34, PyTorch Lightning \cite{Deng2009ImageNetAL,Falcon_PyTorch_Lightning_2019,He2016DeepRL,Paszke_PyTorch_An_Imperative_2019,rw2019timm}\\
Optimization & balanced sampling, Adam optimizer, CE loss, early stopping \cite{Johnson2019SurveyOD,Kingma2015AdamAM}\\
Augmentation & resize (256x256), random cropping (224x224), horizontal flipping \cite{info11020125}\\
Learning rate & initial learning rate search, reduce on plateau \cite{Smith2015CyclicalLR}\\
\bottomrule
\end{tabular}
\end{table}

\end{document}